\documentclass[letterpaper, 10 pt, conference]{ieeeconf}  % Comment this line out if you need a4paper

\IEEEoverridecommandlockouts                              % This command is only needed if 
\usepackage{cite}
\usepackage{graphicx} % for pdf, bitmapped graphics files
\usepackage{siunitx}
\usepackage{subcaption}
\usepackage{xcolor}
\usepackage{tikz}
\usetikzlibrary{shapes.geometric}
\usepackage{comment}
\usepackage{textcomp}
\usepackage{amsmath} % assumes amsmath package installed
\usepackage{amssymb}  % assumes amsmath package installed
\usepackage{todonotes}
\usepackage[breaklinks=true]{hyperref}
\usepackage{gensymb}
\usepackage{algpseudocode}
\usepackage[ruled,vlined]{algorithm2e}
\definecolor{TUMBlue}{rgb}{0.0, 0.40, 0.74}
\definecolor{TUMGray1}{rgb}{0.2, 0.2, 0.2}
\definecolor{TUMGray3}{rgb}{0.8, 0.8, 0.8}
\definecolor{TUMBlue4}{rgb}{0.60, 0.78, 0.92}
\definecolor{TUMIvory}{rgb}{0.85, 0.84, 0.80}
\definecolor{TUMOrange}{rgb}{0.89, 0.45, 0.13}
\definecolor{TUMGreen}{rgb}{0.64, 0.68, 0.0}
\definecolor{TUMGreenWeb}{rgb}{0.62, 0.73, 0.21}
\definecolor{TUMRedWeb}{rgb}{0.92, 0.45, 0.22}
\usepackage{tikz}
\usepackage{textcomp}
\usepackage{lipsum}
\usepackage{fancyhdr} %load fancyhdr package
\newcommand\copyrighttext{%
	\footnotesize \textcopyright~2024 IEEE.  Personal use of this material is permitted.  Permission from IEEE must be obtained for all other uses, in any current or future media, including reprinting/republishing this material for advertising or promotional purposes, creating new collective works, for resale or redistribution to servers or lists, or reuse of any copyrighted component of this work in other works. DOI: 10.1109/MFI62651.2024.10705773
}
\newcommand\copyrightnotice{%
	\begin{tikzpicture}[remember picture,overlay]
		\node[anchor=south,yshift=4pt] at (current page.south) {\fbox{\parbox{\dimexpr\textwidth-\fboxsep-\fboxrule\relax}{\copyrighttext}}};
	\end{tikzpicture}%
}

\title{\LARGE \bf
Multi-LiCa: A Motion- and Targetless Multi - LiDAR-to-LiDAR Calibration Framework}

\author{Dominik Kulmer$^{1}$, Ilir Tahiraj$^{1}$, Andrii Chumak$^{1}$, and Markus Lienkamp$^{1}$% <-this % stops a space
\thanks{The research was partially funded by the Bavarian Research Foundation (BFS) and through basic research funds from the Institute of Automotive Technology.}% <-this % stops a space
\thanks{$^{1}$Dominik Kulmer, Ilir Tahiraj, and Markus Lienkamp are with the Institute of Automotive Technology, Munich Institute of Robotics and Machine Intelligence (MIRMI),
        Technical University of Munich, 85748 Garching, Germany}%
\thanks{Corresponding author: \tt\small dominik.kulmer@tum.de}%
}

\begin{document}
\maketitle
\thispagestyle{empty} % No header or footer on the title page
\pagestyle{empty} % No header or footer globally

% Now start the fancy header style for the next pages
\pagestyle{fancy} 
\fancyhead{} % Clear all header fields
\renewcommand{\headrulewidth}{0pt}  % No decorative line under the header

% Define the header text for the first page after the title page
\fancyhead[L]{2024 IEEE International Conference on Multisensor Fusion and Integration for Intelligent Systems (MFI)}

\copyrightnotice
\thispagestyle{empty} % No header or footer on the title page
\pagestyle{empty} % No header or footer globally

%%%%%%%%%%%%%%%%%%%%%%%%%%%%%%%%%%%%%%%%%%%%%%%%%%%%%%%%%%%%%%%%%%%%%%%%%%%%%%%%
\begin{abstract}
\thispagestyle{fancy}
Today's autonomous vehicles rely on a multitude of sensors to perceive their environment. To improve the perception or create redundancy, the sensor's alignment relative to each other must be known. 
With \textit{Multi-LiCa}, we present a novel approach for the alignment, e.g. calibration. We present an automatic motion- and targetless approach for the extrinsic multi LiDAR-to-LiDAR calibration without the need for additional sensor modalities or an initial transformation input. 
We propose a two-step process with feature-based matching for the coarse alignment and a GICP-based fine registration in combination with a cost-based matching strategy. Our approach can be applied to any number of sensors and positions if there is a partial overlap between the field of view of single sensors. We show that our pipeline is better generalized to different sensor setups and scenarios and is on par or better in calibration accuracy than existing approaches. 
The presented framework is integrated in \textit{ROS 2} but can also be used as a standalone application.
% hier noch: wie viele Datensätze und wie großer Fehler
To build upon our work, our source code is available at \url{https://github.com/TUMFTM/Multi_LiCa}.
\end{abstract}
%
%%%%%%%%%%%%%%%%%%%%%%%%%%%%%%%%%%%%%%%%%%%%%%%%%%%%%%%%%%%%%%%%%%%%%%%%%%%%%%%%
%%%%%%%%%%%%%%%%%%%%%%%%%%%%%%%%%
%% Introduction
%%%%%%%%%%%%%%%%%%%%%%%%%%%%%%%%%
\section{Introduction}
\label{sec:Introduction}%
As autonomous vehicles progress, the necessity for highly accurate and reliable perception systems becomes critical to ensuring safety and guaranteeing high performance in all scenarios of the design domain~\cite{9190036}. LiDAR sensors are integral to environmental perception, particularly for object detection or localization in Global Navigation Satellite System (GNSS)-denied environments. An analysis of autonomous shuttles by Hafemann et al.~\cite{Hafemann.2023} shows a significant variation in the number and positioning of these sensors. 
In addition to the placement, there are also differences due to the different sensor technologies. So-called rotating LiDARs can cover a large field of view (FOV) of up to \SI{360}{\degree}. In contrast, solid-state LiDARs operate in a limited FOV but generally provide a higher point density at further distances. 
This diversity in sensor arrangements is not just evident in autonomous shuttle services but can also be exemplified through datasets on autonomous driving. Liu et al.~\cite{liu2024survey} provide a comprehensive survey on 265 autonomous driving datasets. However, the five most impactful LiDAR perception datasets, with the exception of~\cite{wang2019apolloscape}, show a clear trend toward horizontal positioning of at least one LiDAR with \SI{360}{\degree} FOV on the vehicle's roof~\cite{nuscenes, waymo2020, Geiger2012, Argoverse2}. In contrast to the trend, the Technical University of Munich~(TUM) research vehicle for autonomous driving, EDGAR (Excellent Driving Garching), is equipped with an angled multi-LiDAR setup distributed around the roof frame~\cite{karle2023edgar}. 

\begin{figure}[t]
    \centering
    \includegraphics[angle=-0, width=1.0\linewidth, trim={0cm 0cm 0cm 0cm}, clip]{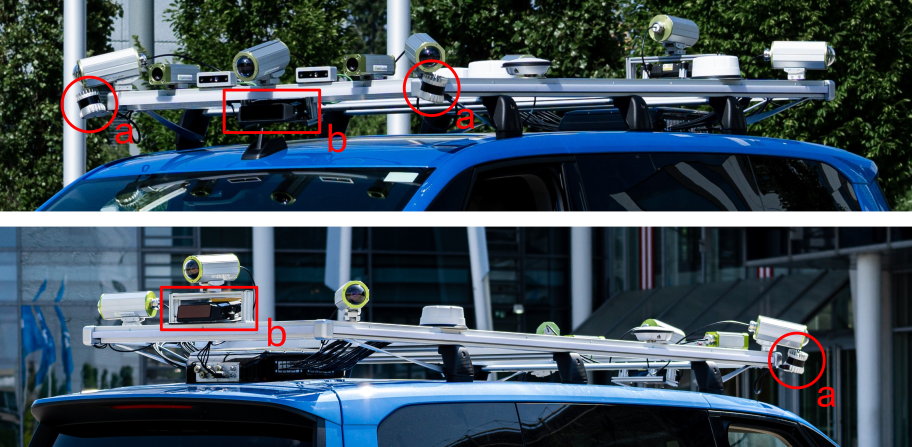}
    \caption{EDGAR LiDAR setup from the front (top image) and back (bottom image) with a) mid-range Ouster OS-1 and b) long-range Seyond Falcon.}
    \label{fig:sensor_placement}
\end{figure}

\begin{figure}[t]
    \centering
    \includegraphics[angle=-0, width=0.8\linewidth, trim={0cm 0cm 0cm 0cm}, clip]{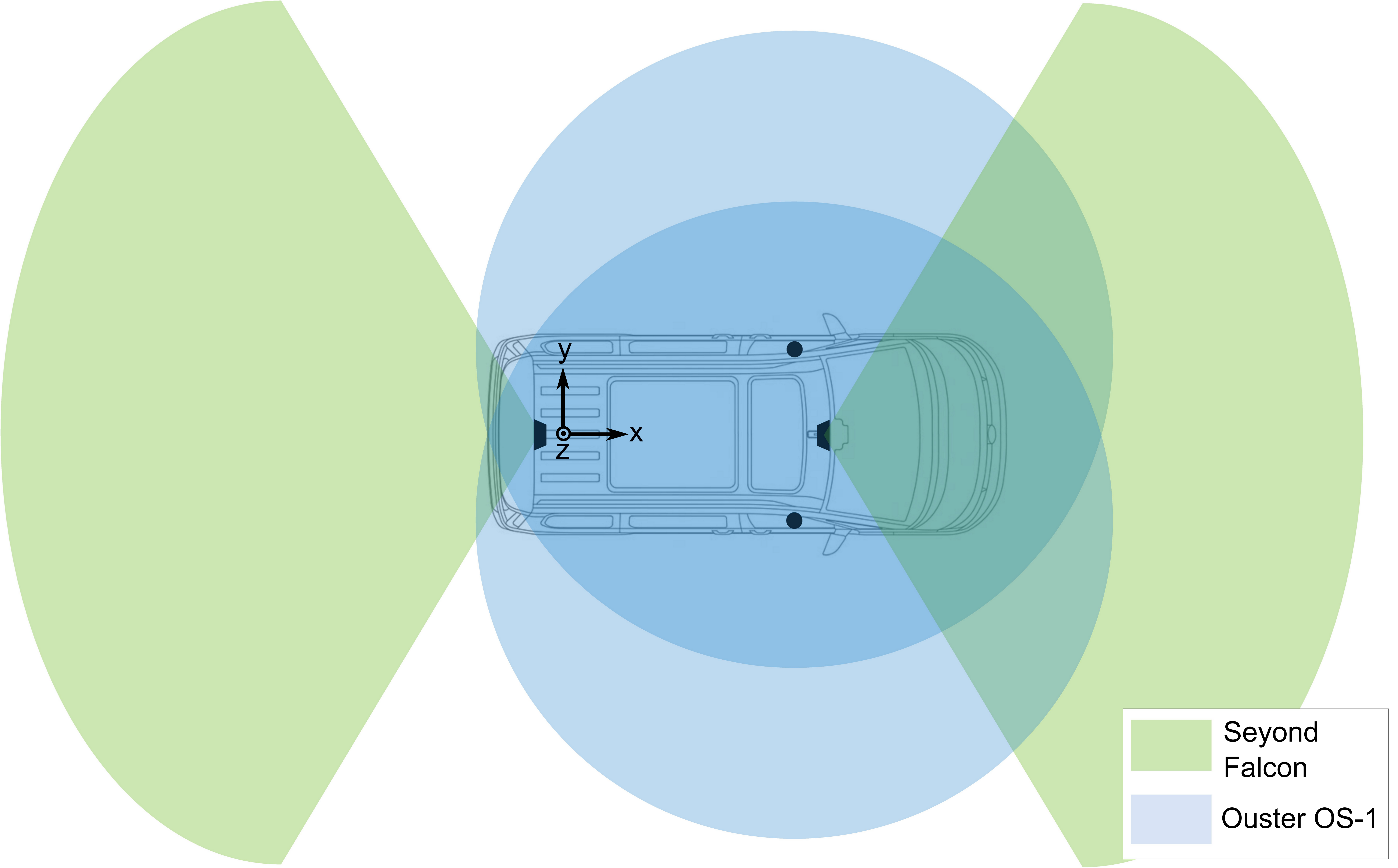}
    \caption{FOV of EDGAR LiDAR sensors.}
    \label{fig:sensor_overlap}
\end{figure}

The distinctive LiDAR setup includes two rotating mid-range Ouster OS-1 and two solid-state long-range Seyond (former Innovusion) Falcon LiDARs (Fig.~\ref{fig:sensor_placement}). Unlike the above introduced autonomous driving datasets~\cite{nuscenes, waymo2020, Geiger2012, Argoverse2}, all four LiDARs on EDGAR are slightly angled to the ground plane. The LiDARs are strategically positioned to enable a \SI{360}{\degree} FOV and a minimized blind spot around the vehicle. 
The arrangement requires a calibration method that makes it possible to calibrate all LiDARs with each other, even if they do not have a common FOV (Fig.~\ref{fig:sensor_overlap}). \\

This work focuses on LiDAR-to-LiDAR calibration methods without the need for other sensor modalities like cameras or IMUs, which can introduce additional complexities and potential inaccuracies. 
Existing multi-LiDAR-to-LiDAR methods often assume a shared FOV overlap with one \textit{main}, e.g. rooftop, LiDAR for the feature matching~\cite{Wei.2022}, posing challenges for EDGAR’s setup due to the lack of overlap between all LiDARs. Motivated by the lack of open-source contributions, we propose a calibration framework that can handle atypical setups like EDGARs without requiring FOV overlap between all LiDARs or reliance on other external sensors. 

The main contributions of this paper are as follows:
\begin{itemize}
    \item We propose an automated, motion- and targetless multi-LiDAR-to-LiDAR calibration framework (Multi-LiCa) that is robust against different LiDAR alignments, sensor types, and calibration scenarios without any restrictions on the relative pose of the sensors other than an overlapping area between single LiDAR FOVs.
    \item We propose a two-step algorithm with a feature-based coarse alignment and a Generalized Iterative Closest Point (GICP)-based fine registration that eliminates the need for initial guesses of the sensor poses and simultaneously enables high calibration accuracy.
    \item We introduce a matching strategy based on point correspondences to determine the overlap of point clouds and formulate a matching and merging strategy to maximize the likelihood of convergence.
    \item We additionally introduce a LiDAR-to-Ground calibration method that can compute $roll$, $pitch$, and $z$ transformation components between a LiDAR and a system's base frame following ISO 8855, which partially automates and simplifies the overall to-Base calibration process.
\end{itemize}

%
%%%%%%%%%%%%%%%%%%%%%%%%%%%%%%%%%
%% Related Work
%%%%%%%%%%%%%%%%%%%%%%%%%%%%%%%%%
\section{Related Work}
\label{sec:RelatedWork}%
LiDAR extrinsic calibration is a well-researched topic with a wide range of methods available. One way to distinguish between the methods is to classify them into \textit{target-based} and \textit{targetless.}

The main principle of target-based calibration is that it uses prior knowledge about so-called targets. This information makes it more accurate than targetless methods~\cite{tsai2021optimising, PERSIC2019217}. However, this method requires calibration targets such as checkerboards that sensors can easily detect, therefore, introducing additional manual effort and reducing the method's usability, especially for setups with a multitude of sensors. \\
\indent Gao and Spletzer~\cite{5509880} showed an automatic calibration approach for multiple LiDARs using retro-reflective targets.  Beltran et al.~\cite{Beltran_2022} introduced a novel calibration target and method for the alignment of different sensor modalities, i.e., cameras, LiDARs, and sensor types such as low and high-resolution LiDARs. Tahiraj et al.~\cite{tahiraj2024gmmcalib} proposed a probabilistic method based on a Gaussian Mixture Model (GMM) for shape fitting to overcome the limitations of point-based matching. A collection of different approaches was introduced with the open-source framework \textit{OpenCalib}~\cite{Yan.5272022}. Sensor calibration is not just an active field in research but also in the industry, with commercial tools such as~\cite{deepenai2024}.\\
\\
Targetless methods aim to solve the practicality issue of the target-based method by trying to find features in the environment and to match them~\cite{Yan.5272022}. The methods can be split into two further categories: \textit{motion-based} and \textit{motionless}.\\
\indent Motion-based methods can rely on external sensors, such as GNSS and/or Inertial Measurement Unit (IMU)~\cite{10081452}, or solely on LiDAR measurements~\cite{liu2021fov, 9811704}.\\
\indent AutoCore~\cite{autocore-ai2023} proposed \textit{LL-Calib}, a ROS~2 calibration toolkit integrated into \textit{Autoware}. One of the relevant methods is LiDAR-to-LiDAR calibration, which builds a local map of one stepwise moved LiDAR to which other LiDARs can be matched. It uses an optimized version of generalized-ICP~\cite{Segal2009} called fast-GICP~\cite{koide2021b} for the scan registration.
Miguel et al.~\cite{10107733} introduced an approach for pairwise LiDAR calibration without any manual initial guess through map-based matching technique.
Das et al.\cite{Das_2023} proposed a strategy of matching absolute GNSS data with estimated LiDAR poses with a novel observability criteria to subsample the poses to be matched for online calibration.
Liu et al.~\cite{9779777} presented \textit{mlcc}, an approach targeted for small FOV LiDARs. They used an adaptive voxelization technique for the feature point matching and minimize the pose estimation in a LiDAR bundle adjustment problem. 
Jiao et al.~\cite{JianhaoJiao.2019b} published a hybrid framework that combines motionless with motion-based methods for calibrating a combination of horizontally positioned LiDARs.\\
\indent In contrast, motionless methods do not require the system to move to perform a calibration, and movement is often not even desired. In addition, the static observation method eliminates the need for time synchronization between sensors, undistortion of the point clouds, and removal of dynamic objects to name a few.\\
\indent Jiao et al.~\cite{JianhaoJiao.2019} demonstrated a method for calibrating a horizontally mounted dual-LiDAR system that relies on a RANSAC-based model for extracting three linearly independent surfaces. Lee et al.~\cite{lee2022planar} introduced the idea of matching planar objects from the different FOVs and could generate more accurate calibrations compared to conventional point cloud registration on their tests.
Ridecell~\cite{ridecell-autoware} developed \textit{Multi LiDAR Calibrator}, a ROS~1 calibration pipeline that relies on the Normal Distribution Transform~(NDT)~\cite{magnusson2007} algorithm. It performs the calibration by matching the relative pose between correspondences in LiDAR scans.
He et al.~\cite{6696597} proposed a framework for the calibration of multiple 2D LiDARs with a multi-type geometric feature matching strategy.
Kim and Kim~\cite{9448104} developed an approach with plane feature matching of structured environments using RANSAC and a nonlinear optimization formulation. 
Yan et al.~\cite{Yan.5272022} introduced \textit{OpenCalib}, a toolbox of different calibration methods, including LiDAR-to-LiDAR calibration developed by Wei et al.~\cite{Wei.2022} named \textit{CROON}. The approach consists of two stages: A rough calibration using the ground plane feature and an Iterative Closest Point~(ICP) with octree optimization refinement. The method was used to calibrate a system of one \textit{main} LiDAR placed on the top of the tested vehicle and four LiDARs positioned at each side.

\subsection{Evaluation of Existing Methods}
We assessed numerous open-source extrinsic LiDAR-to-LiDAR calibration methods, including ~\textit{Multi LiDAR Calibrator}~\cite{ridecell-autoware}, \textit{LL-Calib}~\cite{autocore-ai2023}, \textit{TFAC}~\cite{gong2018target}, ~\textit{CROON}~\cite{Wei.2022} and \textit{mlcc}~\cite{9779777}. 

\begin{figure}[t!]
    \centering
    \includegraphics[angle=-0, width=0.9\linewidth, trim={0cm 0cm 0cm 0cm}, clip]{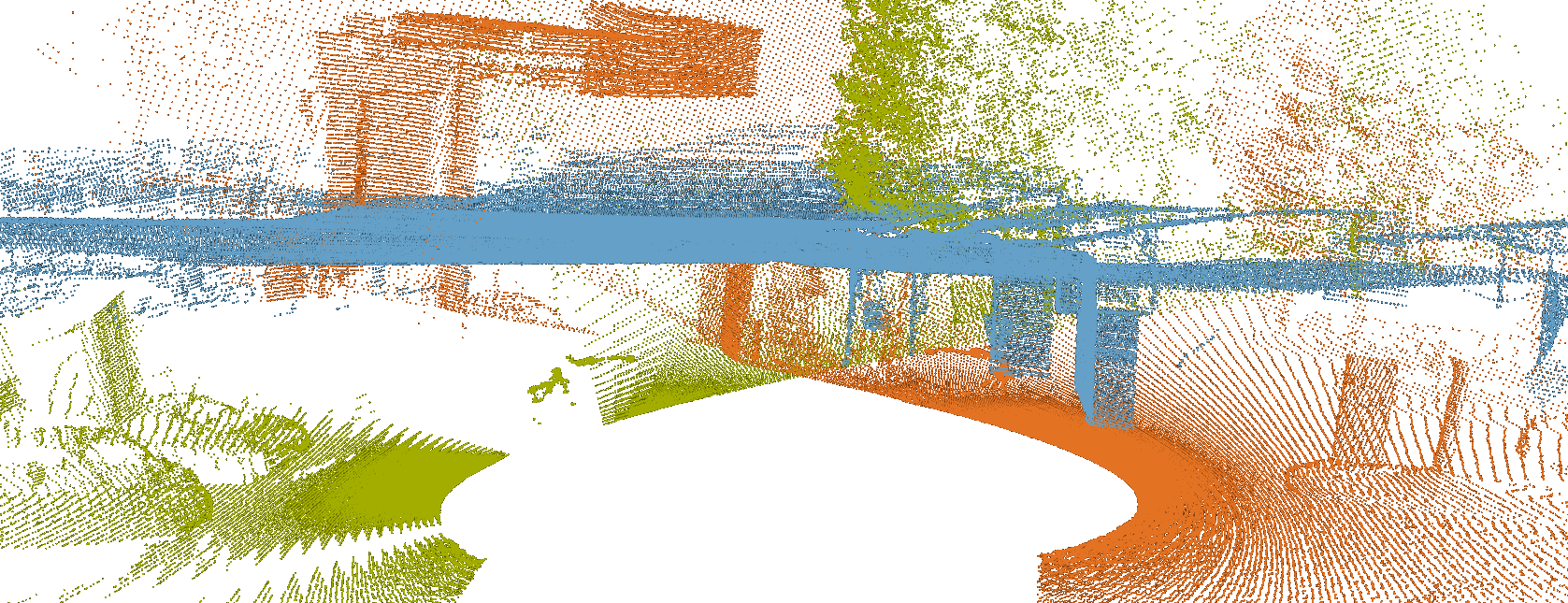}
    \caption{EDGAR front (blue), left (green), and right (orange) LiDAR calibration result with OpenCalibs CROON.}
    \label{fig:opencalib_failure}
\end{figure}

Unfortunately, many repositories are missing any documentation on how to install or use their framework. Others, like \textit{TFAC}, have documentation, but we were not able to compile the code with its requirements. 
\textit{CROON} showed the most promising suit on features with a multi LiDAR-to-LiDAR calibration algorithm and promising results on their published point clouds. Despite the first impression, we were not able to calibrate EDGARs LiDARs, as demonstrated in Fig.~\ref{fig:opencalib_failure}, where the ground was clearly not matched accordingly between the single point clouds.
%

%%%%%%%%%%%%%%%%%%%%%%%%%%%%%%%%%
%% Methodology
%%%%%%%%%%%%%%%%%%%%%%%%%%%%%%%%%
\section{Methodology}
\label{sec:Methodology}%
Our proposed framework can be split into two stages. First the coarse alignment of the point clouds is calculated with a feature-based matching algorithm. The calculated transformation matrices are then used in the second stage, the fine registration, as initial guesses.
\subsection{Coarse Alignment}

To eliminate the need for manually defined initial guesses, we calculate feature vectors of the individual point clouds using Fast Point Feature Histograms (FPFH)~\cite{5152473}. To speed up the process and generate a more even distributed point cloud, we voxelize each cloud for coarse alignment. We use a voxel size of 0.35~m. For each voxel, the containing points are represented by the centroid of the voxel. Then, we define a radius around each point and compute the FPFH feature vector with the nearest neighbors within the defined radius of five times the voxel size. An iterative process was used to optimize both the voxel size and the radius size.\\
We then compute the transformations between the individual source and defined target cloud with TEASER++~\cite{Yang20tro-teaser}. We found that matching within FPFH and RANSAC-based matching is less robust and, therefore, produces more false alignments for varying scenarios and sensor configurations. All calculated transformations are then stored as initial guesses for the following fine registration.

\subsection{Fine Registration}

\begin{figure}[t!]
    \centering
    \includegraphics[angle=-0, width=0.4\linewidth, trim={0cm 0cm 0cm 0cm}, clip]{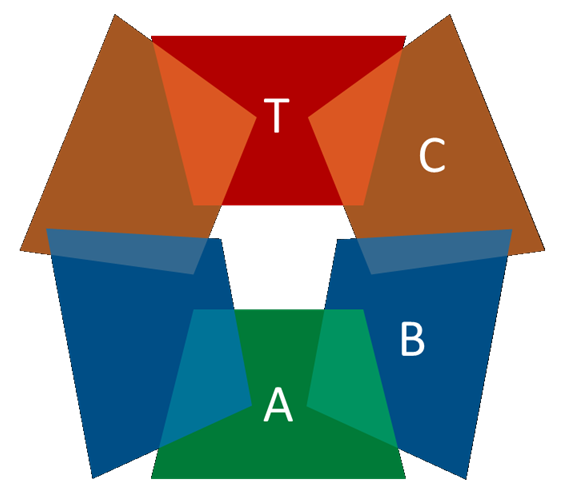}
    \caption{Schematic representation of LiDAR sensors FOV and their overlap.}
    \label{fig:overlap}
\end{figure}

The main algorithm behind the fine registration is the Generalized Iterative Closest Point (GICP)~\cite{Segal2009}.
The basic idea of the multi LiDAR-to-LiDAR calibration is to apply the single LiDAR-to-LiDAR calibration to every LiDAR that has to be calibrated. 
Overlap between sensor data is crucial for the GICP to be able to perform calibration as it tries to match common features that can only be located inside the overlap. However, when trying to calibrate multiple LiDARs to a target LiDAR that does not have a \SI{360}{\degree} FOV, it is possible that some LiDARs do not have enough or any FOV overlap with the target but may have overlap with other LiDARs. Not enough overlap between two point clouds can be detected by looking at the resulting fitness score computed by Open3Ds~\cite{o3d_gicp_param} GICP.
The fitness score measures the overlapping area as the number of inlier correspondences divided by the total points in the target cloud.\\
For this project, a critical assumption is that each LiDAR has either a directly overlapping FOV with the target LiDAR FOV or has overlap with other LiDAR(s) with overlap to the target. This can be cascading dependency to the target, e.g., for LiDARs \textit{T} (target), and \textit{A}, \textit{B}, \textit{C} with overlaps in (\textit{A}, \textit{B}), (\textit{B}, \textit{C}), (\textit{C}, \textit{T}). \textit{A} has overlap with \textit{T} by having overlap with \textit{B} that has overlap to \textit{C} that has overlap with \textit{T} (\(A \rightarrow B \rightarrow C \rightarrow T\)), as can be seen in Fig.~\ref{fig:overlap}.
If the assumption does not hold, at least one LiDAR will not be calibrated to the target LiDAR.
We perform GICP on every source-target LiDAR pair. For every successful calibration, a source point cloud is transformed into the target point cloud and merged, building a larger cloud. The success of an alignment is measured by the fitness score. We found a score of greater~0.2 sufficient. The merge of clouds potentially creates overlapping regions for the remaining LiDARs that initially did not directly overlap with the target LiDAR. The remaining LiDARs are then calibrated to the merged point cloud instead of the initial target cloud (see Algorithm~\ref{alg:fine}). This iterative approach enables a higher likelihood for matching environment features and enables in the first place a calibration of sensor setups, where there is no direct FOV overlap between the desired source and target LiDAR. \\

\begin{algorithm}[httpb]
\SetAlgoLined
\KwResult{Dictionary of calibration names and corresponding calibrations of LiDAR pairs}
 Initialize calibrations as an empty dict\;
 Initialize calibrations\_tmp as an empty list\;
 Initialize not\_calibrated as a set of all LiDARs\;
 \For{each pair of different LiDARs}{
  Compute GICP transformation and store the calibration in calibrations\_tmp\;
 }
 \While{$not\_calibrated \neq \{target$\}}{
  Find the calibration with the highest fitness score\;
  \eIf{highest fitness score is below the threshold}{
   Terminate. The calibration is not possible within the given parameters\;
   }{
   \If{source LiDAR is the target LiDAR}{
    Swap source and target LiDARs\;
    Invert the transformations\;
   }
   Apply the transformation to the source point cloud\;
   Merge the transformed source point cloud with the target point cloud\;
   \If{source transformed to the target}{
    Remove source LiDAR from not\_calibrated\;
    Store {calibration\_name, calibration} pair to the calibrations\;
   }
   Remove all calibrations involving the source LiDAR\;
   \For{each calibration in calibrations\_tmp involving the last calibrated source}{
    Recompute the GICP transformations\;
   }
  }
 }
 \Return{calibrations}\;
 \caption{Fine Registration Method}
 \label{alg:fine}
\end{algorithm}

\subsection{LiDAR-to-Ground Calibration}
Additionally to the LiDAR-to-LiDAR calibration, we propose a simple approach to transform the calculated poses relative to the ground or the base, if certain assumptions are met.
Assuming a stationary and flat ground, a transformation between a flat point cloud approximating the ground and a LiDAR point cloud can be computed via GICP. As there are no distinct features in the generated plane, no $x$, $y$ and $yaw$ calibration can be computed. Nevertheless, the $z$ distance, as well as $roll$ and $pitch$ angle can be determined. \\
The $pitch$ and $roll$ angle of the LiDAR scan can be also computed by RANSAC plane segmentation. Testing showed that the RANSAC approach is more accurate than GICP. \\
Given a plane model: $ax + by + cz + d = 0$ determined by RANSAC, the $pitch$ angle can be computed as
\begin{equation}
    \beta = \text{atan2}(a, \sqrt{b^2 + c^2}) .
\end{equation}
$Roll$ can be calculated similarly.
This concept was inspired by the research conducted by Wei et al.~\cite{Wei.2022}.
\subsection{Parameter Optimization}
For the used methods, e.g., RANSAC, FPFH, TEASER++, and GICP, parameters were examined to determine approximately the optimal values for the calibration process. The priority was on a robust calibration with the best possible accuracy, giving the runtime a lower priority. We computed the variance and the mean average error (MAE) of translational and rotational components and determined the single parameter set used for all calibrations across all configurations and datasets used in this work. These can be taken from our GitHub repository due to their number.
\subsection{Dataset Preparation}

\begin{figure*}
  \includegraphics[width=\textwidth,height=5.5cm]{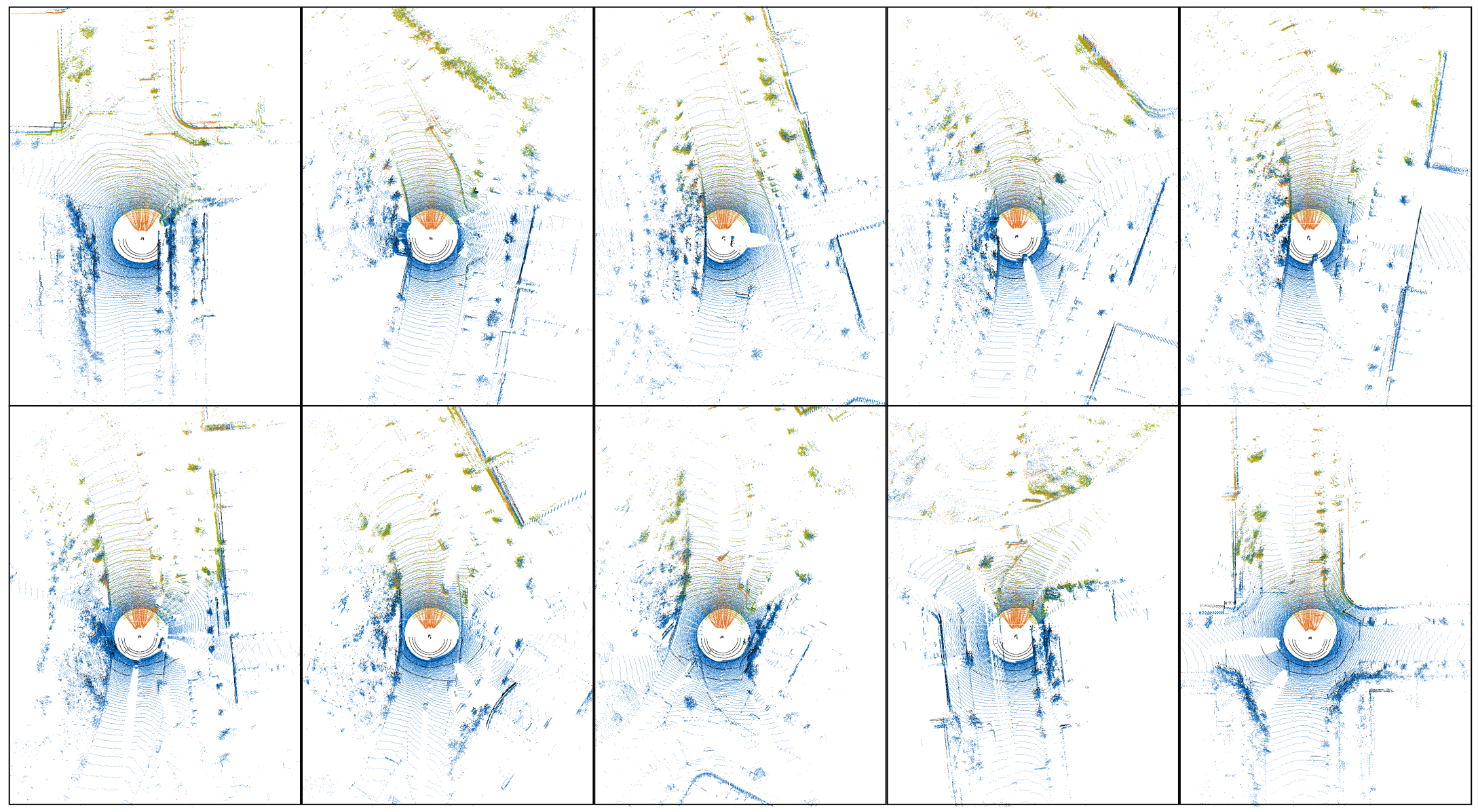}
  \caption{Samples of the HeLiPR dataset with LiDAR point clouds of Ouster (blue), Velodyne (black), Livox (orange) and Aeva (green).}
  \label{fig:raw_helipr}
\end{figure*}

\begin{figure*}
  \includegraphics[width=\textwidth,height=6.5cm]{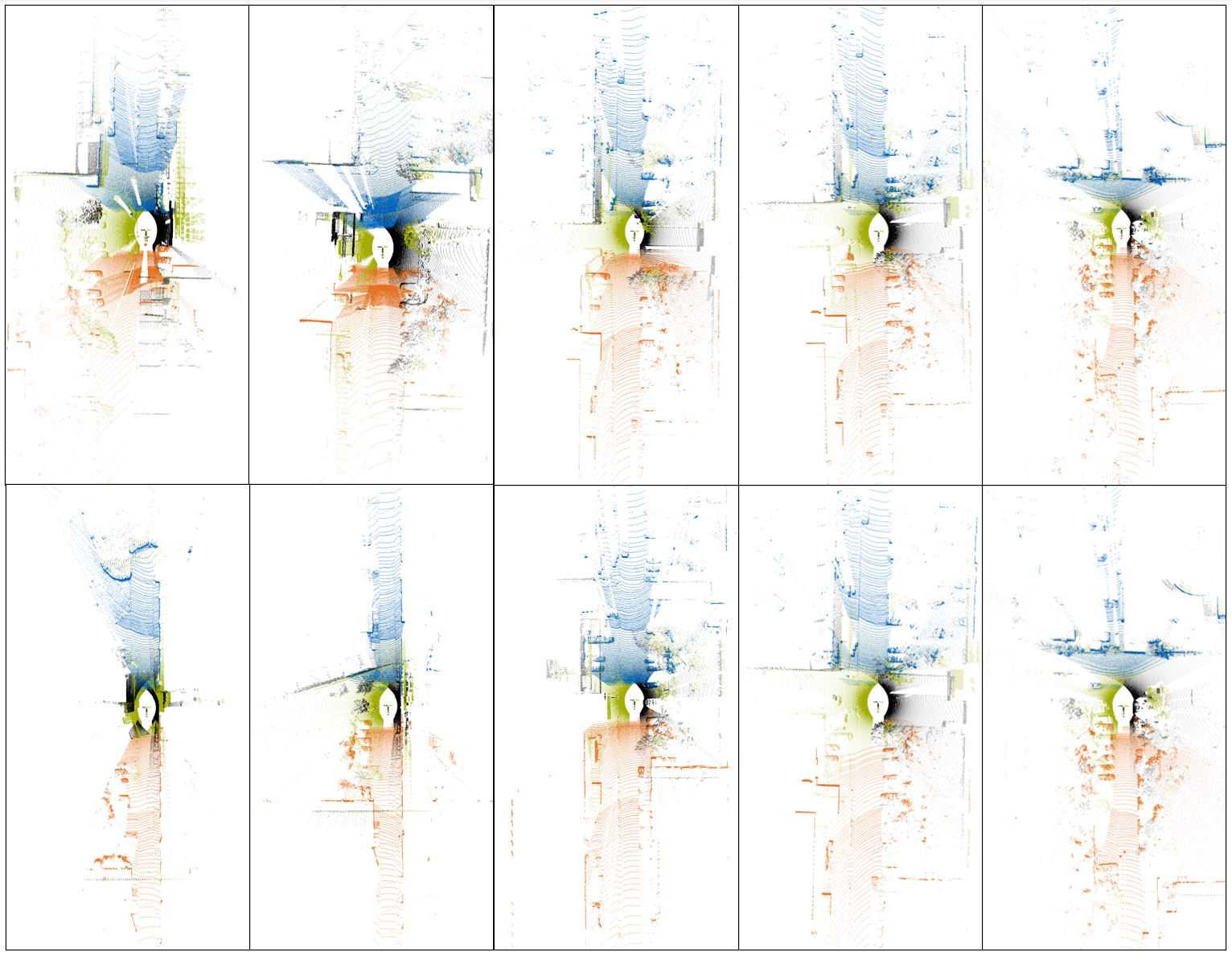}
  \caption{Samples of the EDGAR dataset with front (blue), rear (orange), left (green) and right (black) LiDAR point clouds.}
  \label{fig:raw_edgar}
\end{figure*}

For the evaluation of the proposed method, we choose ten samples of EDGAR data (Fig. ~\ref{fig:raw_edgar}) and ten samples of the HeLiPR~\cite{jung2023helipr} sample dataset (Fig. ~\ref{fig:raw_helipr}). HeLiPR provides data of an Ouster OS2-128, Velodyne VLP-16, Livox Avia, and Aeva Aeries II. This poses an interesting challenge for multi-LiDAR calibration, as the dataset is comprised of two rotating and two solid-state LiDARs. Besides, the type, the scan pattern, and resolution vary widely. HeLiPR provides ground truth data in the form of absolute poses for each LiDAR. The raw dataset clouds are distorted, therefore, we used the provided \textit{toolbox}\footnote{https://github.com/minwoo0611/HeLiPR-Pointcloud-Toolbox} to correct the point clouds.
For the EDGAR data, we sampled ten scenes from recorded test drives. To eliminate the need for time synchronization and the already above mentioned further disadvantages of motion-based data, we only recorded static data. The ground truth for EDGAR was calculated with a manual calibration process. We placed a multitude of cubes with a side length of 0.5~m around the vehicle, focusing on the overlapping zones of the individual LiDAR scans. Then, we manually aligned the visual edges using the ROS~2 tool RVIZ2. 
\subsection{Evaluation Metrics}
We calculated the poses and the deviations from the ground truth for both datasets individually for each LiDAR and scene. We display the translational error as the Root Mean Square Error (RMSE) for each individual axis. These values represent a combination of all scenes and LiDARs.\\
For the rotational error, we also compute the RMSE based on the individual axis. The individual rotation error between two orientations is represented by the angle of the relative rotation needed to align one orientation with the other. 
%%%%%%%%%%%%%%%%%%%%%%%%%%%%%%%%%
%% Results
%%%%%%%%%%%%%%%%%%%%%%%%%%%%%%%%%
\section{Results}
\begin{figure*}
  \includegraphics[width=\textwidth,height=11.2cm]{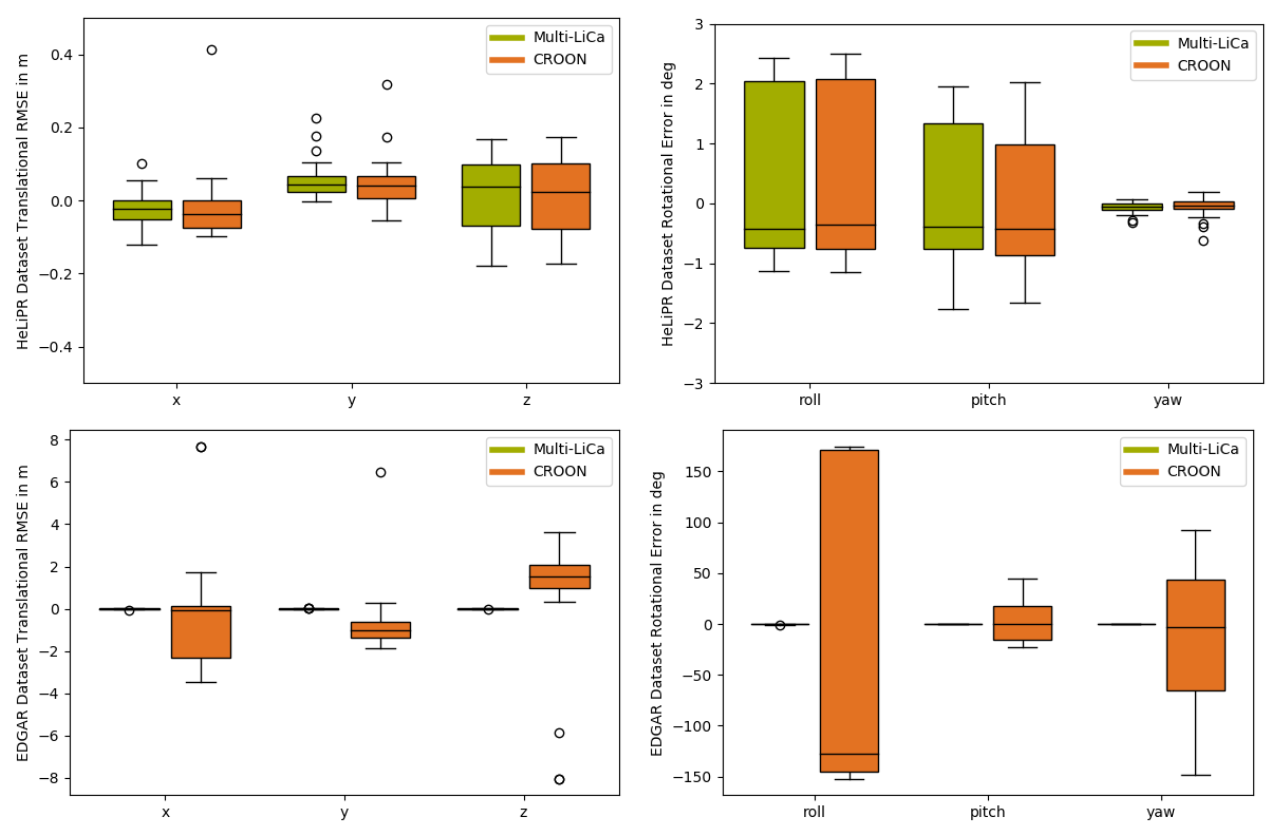}
  \caption{Translational and rotational error of Multi-LiCa and CROON on the HeLiPR and EDGAR dataset.}
  \label{fig:results_new}
\end{figure*}

To evaluate our approach, we computed the calibrations for the above-introduced data for our approach and CROON~\cite{Wei.2022}, as it is the closest comparison regarding the functionality of multi LiDAR-to-LiDAR calibration.

\subsection{HeLiPR Dataset}
\begin{figure}[t!]
    \centering
    \includegraphics[angle=-0, width=0.8\linewidth, trim={0cm 0cm 0cm 0cm}, clip]{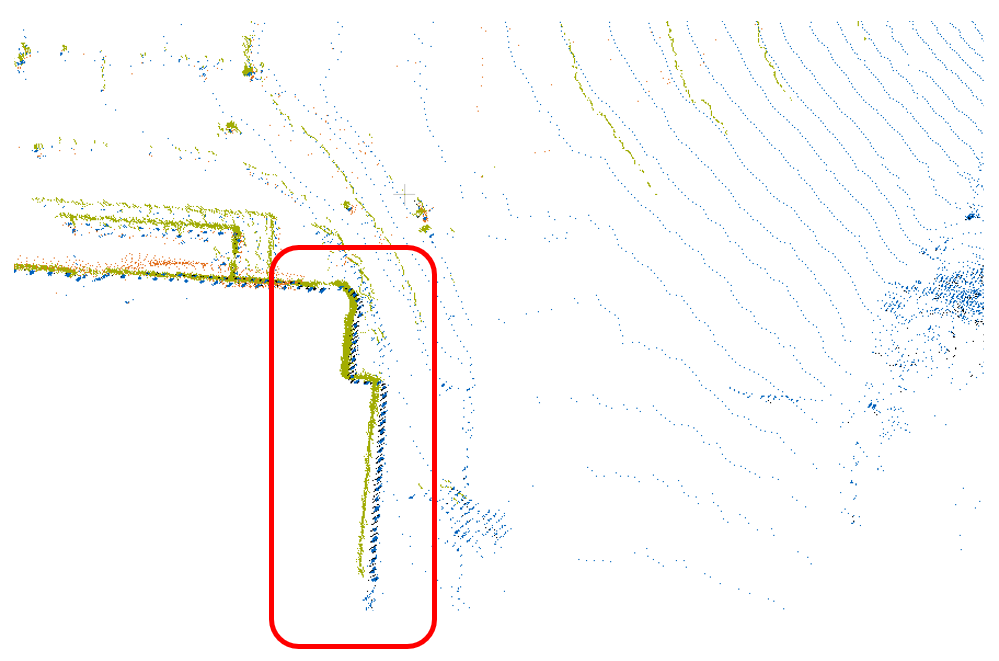}
    \caption{Misalignment of HeLiPR Aeva (green) and Ouster (blue) LiDAR data.}
    \label{fig:helipr_fail}
\end{figure}

Fig.~\ref{fig:results_new} shows the calibration results for our approach and CROON on the sampled HeLiPR dataset. Focusing on the translational error, both methods are comparable in accuracy for the individual axis with the exception that CROON posts more outliers. CROON was for two samples not able to converge the Velodyne LiDAR close to the ground truth. As these values derive much from the rest, they are not visible in the plot. \\
The rotational error again shows a similar trend for both methods. It is apparent that both methods show a large absolute error for $roll$ and $pitch$ but a similar one. For both methods, this deviation from the ground truth is not apparent when visualizing the calibrated clouds. It could hint at an error in the HeLiPR ground truth data. We transformed the first of our sampled data with the ground truth pose and found a substantial deviation in translation and rotation for the Aeva data (Fig.~\ref{fig:helipr_fail}). Therefore, these values have to be looked at in relative terms for the comparison of both algorithms. The average computation time on a single core of an x86 CPU with 3~GHz was 9 seconds for CROON and 3 seconds for Multi-LiCa.
\subsection{EDGAR Dataset}
\begin{figure}[t!]
    \centering
    \includegraphics[angle=-0, width=1\linewidth, trim={0cm 0cm 0cm 0cm}, clip]{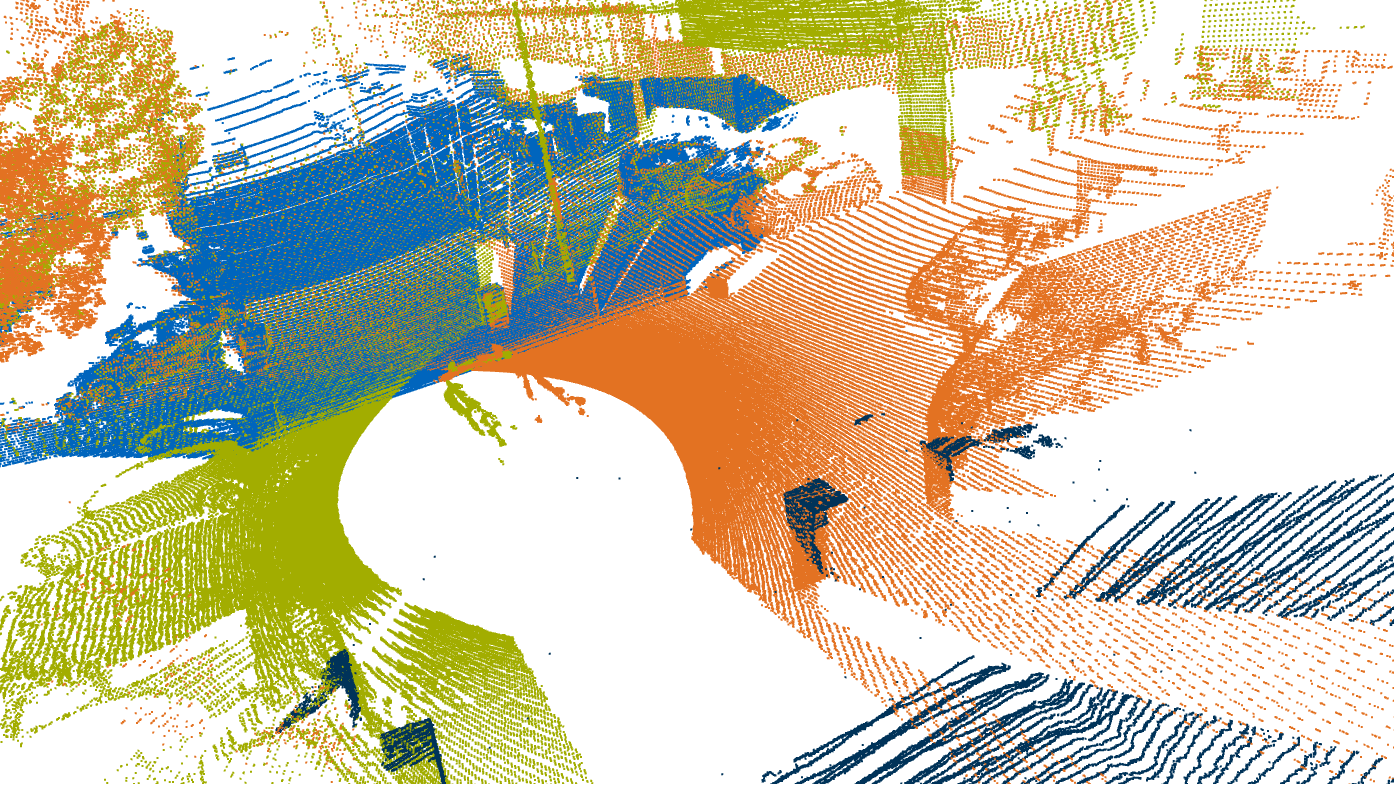}
    \caption{Fused LiDAR point clouds of the front (blue), left (green), right (orange), and rear (black) LiDARs after the sensor calibration of EDGAR with Multi-LiCa.}
    \label{fig:edgar_result}
\end{figure}

For the EDGAR dataset, a completely different trend is visible. While our approach posts translational positions close to the ground truth, CROON fails in many cases to do so. Apparent from the standard deviation and the outliers, CROON struggles to compute a meaningful calibration result. \\
The observed trend is even more pronounced in the rotational error. Analyzing the individual LiDAR calibrations. CROON especially fails to align the rear LiDAR. In the HeLiPR data the difference between the initial alignment and the ground truth is small compared to the EDGAR data. Due to the orientation of EDGAR's LiDARs, a robust initial coarse estimate is particularly important. CROON performs significantly worse here than Multi-LiCa, resulting in high deviations and misalignments. CROON bases the first alignment on the RANSAC algorithm, which we found can lead to non-robust alignments between the clouds for the EDGAR dataset. Reasons could be the small overlapping FOV, the missing of one \textit{main} LiDAR overlapping with all source point clouds, and the orientation differences between the point clouds. Croon also showed very high calculation times. The pose estimation of the first sample took over 1300 seconds, while Multi-LiCa also took an average of 3 seconds. \\
A calibrated alignment of EDGAR's LiDAR sensors with our approach can be seen in Fig.~\ref{fig:edgar_result}. \\
%

%%%%%%%%%%%%%%%%%%%%%%%%%%%%%%%%%
%% Discussion
%%%%%%%%%%%%%%%%%%%%%%%%%%%%%%%%%
\section{Discussion}
\label{sec:Discussion}%
With Multi-LiCa, we showed a robust and accurate point cloud calibration approach that does not need calibration targets or specific environment features for the data matching. 
Multi-LiCa benefits from a robust initial guess computation for the fine registration, posing state-of-the-art calibration results across the tested scenes. \\
We showed that our method is sufficiently stable to handle different sensor configurations, types, and scan patterns without the need for a manually defined initial guess for the pose alignment. 
Tests on data from HeLiPR and EDGAR showed visually and quantitative accurate results compared to existing calibration tools. 
%
%%%%%%%%%%%%%%%%%%%%%%%%%%%%%%%%%
%% Conclusion
%%%%%%%%%%%%%%%%%%%%%%%%%%%%%%%%%
\section{Conclusion and Future Work}
Multi-LiCa presents a novel multi LiDAR-to-LiDAR calibration framework that handles data from different LiDAR setups with non-direct sensor FOV overlap and minimizes user input while maintaining generalizability. It was tested on a total of twenty different scenes of two different sensor setups, yielding satisfactory results and showing potential for broader application. However, additional testing is needed for conclusive proof of generalizability.

Future work should encompass improving the accuracy further through the use of optimization techniques, such as pose graph optimization. Evaluation on a broader set of LiDAR data should be carried out to further evaluate the generalizibility of the method.
% \section*{ACKNOWLEDGMENT}
%
%%%%%%%%%%%%%%%%%%%%%%%%%%%%%%%%%%%%%%%%%%%%%%%%%%%%%%%%%%%%%%%%%%%%%%%%%%%%%%%%
%
\bibliographystyle{IEEEtran}
\bibliography{references}

\begin{thebibliography}{10}
\providecommand{\url}[1]{#1}
\csname url@rmstyle\endcsname
\providecommand{\newblock}{\relax}
\providecommand{\bibinfo}[2]{#2}
\providecommand\BIBentrySTDinterwordspacing{\spaceskip=0pt\relax}
\providecommand\BIBentryALTinterwordstretchfactor{4}
\providecommand\BIBentryALTinterwordspacing{\spaceskip=\fontdimen2\font plus
\BIBentryALTinterwordstretchfactor\fontdimen3\font minus
  \fontdimen4\font\relax}
\providecommand\BIBforeignlanguage[2]{{%
\expandafter\ifx\csname l@#1\endcsname\relax
\typeout{** WARNING: IEEEtran.bst: No hyphenation pattern has been}%
\typeout{** loaded for the language `#1'. Using the pattern for}%
\typeout{** the default language instead.}%
\else
\language=\csname l@#1\endcsname
\fi
#2}}

\bibitem{9190036}
C.-P. Hsu, B.~Li, B.~Solano-Rivas, A.~R. Gohil, P.~H. Chan, A.~D. Moore, and
  V.~Donzella, ``A review and perspective on optical phased array for
  automotive lidar,'' \emph{IEEE Journal of Selected Topics in Quantum
  Electronics}, vol.~27, no.~1, pp. 1--16, 2021.

\bibitem{Hafemann.2023}
P.~Hafemann, A.~L. Prat, and M.~Lienkamp, ``Positioning and body integration of
  sensors in autonomous shuttle vehicles,'' \emph{Stuttgarter Symposium Fuer
  Produktentwicklun}, 2023.

\bibitem{liu2024survey}
M.~Liu, E.~Yurtsever, J.~Fossaert, X.~Zhou, W.~Zimmer, Y.~Cui, B.~L. Zagar, and
  A.~C. Knoll, ``A survey on autonomous driving datasets: Statistics,
  annotation quality, and a future outlook,'' 2024.

\bibitem{wang2019apolloscape}
P.~Wang, X.~Huang, X.~Cheng, D.~Zhou, Q.~Geng, and R.~Yang, ``The apolloscape
  open dataset for autonomous driving and its application,'' \emph{IEEE
  transactions on pattern analysis and machine intelligence}, 2019.

\bibitem{nuscenes}
H.~Caesar, V.~Bankiti, A.~H. Lang, S.~Vora, V.~E. Liong, Q.~Xu, A.~Krishnan,
  Y.~Pan, G.~Baldan, and O.~Beijbom, ``nuscenes: A multimodal dataset for
  autonomous driving,'' in \emph{CVPR}, 2020.

\bibitem{waymo2020}
P.~Sun, H.~Kretzschmar, X.~Dotiwalla, A.~Chouard, V.~Patnaik, P.~Tsui, J.~Guo,
  Y.~Zhou, Y.~Chai, B.~Caine, V.~Vasudevan, W.~Han, J.~Ngiam, H.~Zhao,
  A.~Timofeev, S.~Ettinger, M.~Krivokon, A.~Gao, A.~Joshi, Y.~Zhang, J.~Shlens,
  Z.~Chen, and D.~Anguelov, ``Scalability in perception for autonomous driving:
  Waymo open dataset,'' in \emph{Proceedings of the IEEE/CVF Conference on
  Computer Vision and Pattern Recognition (CVPR)}, June 2020.

\bibitem{Geiger2012}
A.~Geiger, P.~Lenz, and R.~Urtasun, ``Are we ready for autonomous driving? the
  kitti vision benchmark suite,'' in \emph{2012 IEEE Conference on Computer
  Vision and Pattern Recognition}, 2012, pp. 3354--3361.

\bibitem{Argoverse2}
B.~Wilson, W.~Qi, T.~Agarwal, J.~Lambert, J.~Singh, S.~Khandelwal, B.~Pan,
  R.~Kumar, A.~Hartnett, J.~K. Pontes, D.~Ramanan, P.~Carr, and J.~Hays,
  ``Argoverse 2: Next generation datasets for self-driving perception and
  forecasting,'' in \emph{Proceedings of the Neural Information Processing
  Systems Track on Datasets and Benchmarks (NeurIPS Datasets and Benchmarks
  2021)}, 2021.

\bibitem{karle2023edgar}
P.~Karle, T.~Betz, M.~Bosk, F.~Fent, N.~Gehrke, M.~Geisslinger, L.~Gressenbuch,
  P.~Hafemann, S.~Huber, M.~H{\"u}bner, \emph{et~al.}, ``Edgar: An autonomous
  driving research platform--from feature development to real-world
  application,'' \emph{arXiv preprint arXiv:2309.15492}, 2023.

\bibitem{Wei.2022}
P.~Wei, G.~Yan, Y.~Li, K.~Fang, X.~Cai, J.~Yang, and W.~Liu, ``{CROON:
  Automatic Multi-LiDAR Calibration and Refinement Method in Road Scene},'' in
  \emph{{2022 IEEE/RSJ International Conference on Intelligent Robots and
  Systems (IROS)}}.\hskip 1em plus 0.5em minus 0.4em\relax IEEE, 2022, pp.
  12\,857--12\,863.

\bibitem{tsai2021optimising}
D.~Tsai, S.~Worrall, M.~Shan, A.~Lohr, and E.~Nebot, ``Optimising the selection
  of samples for robust lidar camera calibration,'' 2021.

\bibitem{PERSIC2019217}
J.~Peršić, I.~Marković, and I.~Petrović, ``Extrinsic 6dof calibration of a
  radar–lidar–camera system enhanced by radar cross section estimates
  evaluation,'' \emph{Robotics and Autonomous Systems}, vol. 114, pp. 217--230,
  2019.

\bibitem{5509880}
C.~Gao and J.~R. Spletzer, ``On-line calibration of multiple lidars on a mobile
  vehicle platform,'' in \emph{2010 IEEE International Conference on Robotics
  and Automation}, 2010, pp. 279--284.

\bibitem{Beltran_2022}
\BIBentryALTinterwordspacing
J.~Beltran, C.~Guindel, A.~de~la Escalera, and F.~Garcia, ``Automatic extrinsic
  calibration method for lidar and camera sensor setups,'' \emph{IEEE
  Transactions on Intelligent Transportation Systems}, vol.~23, no.~10, p.
  17677–17689, Oct. 2022. [Online]. Available:
  \url{http://dx.doi.org/10.1109/TITS.2022.3155228}
\BIBentrySTDinterwordspacing

\bibitem{tahiraj2024gmmcalib}
I.~Tahiraj, F.~Fent, P.~Hafemann, E.~Ye, and M.~Lienkamp, ``Gmmcalib: Extrinsic
  calibration of lidar sensors using gmm-based joint registration,'' 2024.

\bibitem{Yan.5272022}
G.~Yan, L.~Zhuochun, C.~Wang, C.~Shi, P.~Wei, X.~Cai, T.~Ma, Z.~Liu, Z.~Zhong,
  Y.~Liu, M.~Zhao, Z.~Ma, and Y.~Li, ``{OpenCalib: A Multi-sensor Calibration
  Toolbox for Autonomous Driving},'' 2022.

\bibitem{deepenai2024}
{Deepen~AI}, ``Lidar-lidar calibration,''
  \url{https://help.deepen.ai/deepen-ai-enterprise/calibration/lidar-lidar-calibration#calibration-instructions-page},
  2024, accessed: Jan. 31, 2024.

\bibitem{10081452}
S.~Das, L.~a. Klinteberg, M.~Fallon, and S.~Chatterjee, ``Observability-aware
  online multi-lidar extrinsic calibration,'' \emph{IEEE Robotics and
  Automation Letters}, vol.~8, no.~5, pp. 2860--2867, 2023.

\bibitem{liu2021fov}
X.~Liu and F.~Zhang, ``Extrinsic calibration of multiple lidars of small fov in
  targetless environments,'' \emph{IEEE Robotics and Automation Letters},
  vol.~6, no.~2, pp. 2036--2043, 2021.

\bibitem{9811704}
S.~Das, N.~Mahabadi, A.~Djikic, C.~Nassir, S.~Chatterjee, and M.~Fallon,
  ``Extrinsic calibration and verification of multiple non-overlapping field of
  view lidar sensors,'' in \emph{2022 International Conference on Robotics and
  Automation (ICRA)}, 2022, pp. 919--925.

\bibitem{autocore-ai2023}
{AutoCore}, ``calibration\_tools,''
  \url{https://github.com/autocore-ai/calibration_tools/tree/main/lidar-lidar-calib},
  2023, accessed: Nov. 13, 2023, GitHub.

\bibitem{Segal2009}
A.~Segal, D.~Haehnel, and S.~Thrun, ``Generalized-icp,'' 2009.

\bibitem{koide2021b}
K.~Koide, M.~Yokozuka, S.~Oishi, and A.~Banno, ``Voxelized gicp for fast and
  accurate 3d point cloud registration,'' in \emph{2021 IEEE International
  Conference on Robotics and Automation (ICRA)}, 2021, pp. 11\,054--11\,059.

\bibitem{10107733}
M.~A. de~Miguel, C.~Guindel, A.~Al-Kaff, and F.~García, ``High-accuracy
  patternless calibration of multiple 3-d lidars for autonomous vehicles,''
  \emph{IEEE Sensors Journal}, vol.~23, no.~11, pp. 12\,200--12\,208, 2023.

\bibitem{Das_2023}
\BIBentryALTinterwordspacing
S.~Das, L.~a. Klinteberg, M.~Fallon, and S.~Chatterjee, ``Observability-aware
  online multi-lidar extrinsic calibration,'' \emph{IEEE Robotics and
  Automation Letters}, vol.~8, no.~5, p. 2860–2867, May 2023. [Online].
  Available: \url{http://dx.doi.org/10.1109/LRA.2023.3262176}
\BIBentrySTDinterwordspacing

\bibitem{9779777}
X.~Liu, C.~Yuan, and F.~Zhang, ``Targetless extrinsic calibration of multiple
  small fov lidars and cameras using adaptive voxelization,'' \emph{IEEE
  Transactions on Instrumentation and Measurement}, vol.~71, pp. 1--12, 2022.

\bibitem{JianhaoJiao.2019b}
J.~Jiao, Y.~Yu, Q.~Liao, H.~Ye, and M.~Liu, ``Automatic calibration of multiple
  3d lidars in urban environments,'' 2019.

\bibitem{JianhaoJiao.2019}
{Jianhao Jiao}, {Qinghai Liao}, {Yilong Zhu}, {Tianyu Liu}, {Yang Yu}, {Rui
  Fan}, {Lujia Wang}, and {Ming Liu}, \emph{A Novel Dual-Lidar Calibration
  Algorithm Using Planar Surfaces}.\hskip 1em plus 0.5em minus 0.4em\relax
  Piscataway, New Jersey: IEEE, 2019.

\bibitem{lee2022planar}
\BIBentryALTinterwordspacing
H.~Lee and W.~Chung, ``Extrinsic calibration of multiple 3d lidar sensors by
  the use of planar objects,'' \emph{Sensors}, vol.~22, no.~19, 2022. [Online].
  Available: \url{https://www.mdpi.com/1424-8220/22/19/7234}
\BIBentrySTDinterwordspacing

\bibitem{ridecell-autoware}
{Ridecell}, ``Autoware ros1 ndt-calibration,''
  \url{https://github.com/Ridecell/Autoware/blob/master/ros/src/sensing/fusion/packages/multi_lidar_calibrator},
  2023, accessed: Nov. 17, 2023, GitHub.

\bibitem{magnusson2007}
M.~Magnusson, A.~Lilienthal, and T.~Duckett, ``Scan registration for autonomous
  mining vehicles using 3d-ndt,'' \emph{Journal of Field Robotics}, vol.~24,
  no.~10, pp. 803--827, 2007.

\bibitem{6696597}
M.~He, H.~Zhao, F.~Davoine, J.~Cui, and H.~Zha, ``Pairwise lidar calibration
  using multi-type 3d geometric features in natural scene,'' in \emph{2013
  IEEE/RSJ International Conference on Intelligent Robots and Systems}, 2013,
  pp. 1828--1835.

\bibitem{9448104}
D.-H. Kim and G.-W. Kim, ``Automatic multiple lidar calibration based on the
  plane features of structured environments,'' \emph{IEEE Access}, vol.~9, pp.
  84\,387--84\,402, 2021.

\bibitem{gong2018target}
Z.~Gong, C.~Wen, C.~Wang, and J.~Li, ``A target-free automatic self-calibration
  approach for multibeam laser scanners,'' \emph{IEEE Transactions on
  Instrumentation and Measurement}, vol.~67, no.~1, pp. 238--240, 2018.

\bibitem{5152473}
R.~B. Rusu, N.~Blodow, and M.~Beetz, ``Fast point feature histograms (fpfh) for
  3d registration,'' in \emph{2009 IEEE International Conference on Robotics
  and Automation}, 2009, pp. 3212--3217.

\bibitem{Yang20tro-teaser}
H.~Yang, J.~Shi, and L.~Carlone, ``{TEASER: Fast and Certifiable Point Cloud
  Registration},'' \emph{{IEEE} Trans. Robotics}, 2020.

\bibitem{o3d_gicp_param}
Open3D, ``Generalized icp registration — open3d 0.17.0 documentation,''
  \url{http://www.open3d.org/docs/latest/python_api/open3d.pipelines.registration.registration_generalized_icp.html},
  2023, accessed: Nov. 29, 2023.

\bibitem{jung2023helipr}
M.~Jung, W.~Yang, D.~Lee, H.~Gil, G.~Kim, and A.~Kim, ``Helipr: Heterogeneous
  lidar dataset for inter-lidar place recognition under spatial and temporal
  variations,'' 2023.

\end{thebibliography}
\end{document}